\documentclass{article}

\usepackage{microtype}
\usepackage{graphicx}
\usepackage{subfigure}
\usepackage{booktabs} 

\usepackage{hyperref}

\usepackage[accepted]{icml2024}

\usepackage{amsmath}
\usepackage{amssymb}
\usepackage{mathtools}
\usepackage{amsthm}

\usepackage[capitalize,noabbrev]{cleveref}

\theoremstyle{plain}

\theoremstyle{definition}

\theoremstyle{remark}

\usepackage[textsize=tiny]{todonotes}

\icmltitlerunning{Learning Spatio-Temporal Patterns of Polar Ice Layers With Physics-Informed Graph Neural Network}

\begin{document}

\twocolumn[
\icmltitle{Learning Spatio-Temporal Patterns of Polar Ice Layers \\With Physics-Informed Graph Neural Network}




\begin{icmlauthorlist}
\icmlauthor{Zesheng Liu}{cse}
\icmlauthor{Maryam Rahnemoonfar}{cse,cee}
\end{icmlauthorlist}

\icmlaffiliation{cse}{Department of Computer Science and Engineering, Lehigh University, Bethlehem, PA, USA}

\icmlaffiliation{cee}{Department of Civil and Environmental Engineering, Lehigh University, Bethlehem, PA, USA}

\icmlcorrespondingauthor{Maryam Rahnemoonfar}{maryam@lehigh.edu}

\icmlkeywords{Machine Learning, ICML}

\vskip 0.3in
]



 \printAffiliationsAndNotice{}  

\begin{abstract}
Learning spatio-temporal patterns of polar ice layers is crucial for monitoring the change in ice sheet balance and evaluating ice dynamic processes. While a few researchers focus on learning ice layer patterns from echogram images captured by airborne snow radar sensors via different convolutional neural networks, the noise in the echogram images proves to be a major obstacle. Instead, we focus on geometric deep learning based on graph neural networks to learn the spatio-temporal patterns from thickness information of shallow ice layers and make predictions for deep layers. In this paper, we propose a physics-informed hybrid graph neural network that combines the GraphSAGE framework for graph feature learning with the long short-term memory (LSTM) structure for learning temporal changes, and introduce measurements of physical ice properties from Model Atmospheric Regional (MAR) weather model as physical node features. We found that our proposed network can consistently outperform the current non-inductive or non-physical model in predicting deep ice layer thickness.

\end{abstract}

\section{Introduction}
As the global temperature keeps rising in recent years, various research has revealed that the mass loss of polar ice sheets has been accelerated\cite{Forsberg2017, Zwally2011, Mouginot2019,Rignot2011}. Ice sheets are composed of several internal ice layers formed in different years. A precise understanding of these internal ice layers can provide valuable information about past snowfall melting and accumulation. This information can further enable a better comprehension of the climate system changes and reduce the uncertainties of future trend prediction.

The traditional method to capture the status of internal ice layers is through onsite ice core\cite{PATERSON1994378}. However, the coverage of the ice core is limited, and the measurements are discrete, making it impossible to understand the continuous change of the internal ice layer. Moreover, onsite ice cores are expensive and time-consuming to obtain, and the drilling process causes destructive damage to the ice sheets. In recent years, airborne snow radar sensor\cite{AirborneRadar} has become a more effective tool to continuously capture the status of internal ice layers. One typical example is the snow radar operated by Center for Remote Sensing of Ice Sheets (CReSIS), as part of NASA's Operation IceBridge Mission\cite{snowradar}. Internal layers with different depths are captured in echogram images by measuring the reflection signal strength\cite{Arnold_2020}, as shown in Figure \ref{fig:dataset} (a). 

Various automatic algorithms \cite{Carrer_2017,Ferro_2013,Panton_2013,Koenig_2016,MacGregor_2015} and neural networks \cite{DeepIceLayerTracking,DeepLearningOnAirborneRadar, Rahnemoonfar_2021_JOG, Yari_2021_JSTAR} has been proposed to detect ice layer boundaries directly from the raw echogram images. However, noise in the raw echogram images has been proven to be a significant issue. To reduce the effect of noise and better predict the thickness of deep ice layers, Zalatan et al. focused on learning spatio-temporal relationships between ice layers formed at different years through graph neural networks\yrcite{Zalatan2023,Zalatan_igarss,zalatan_icip}. As the current state-of-the-art model, they applied an adaptive recurrent graph convolution neural network (AGCN-LSTM) that learns the patterns from a few shallow ice layers and makes predictions for the thickness of deep ice layers. 

Physics-informed machine learning has been rapidly developed in recent years. It is a promising learning framework that leverages machine learning's ability in pattern recognition and physical methods' strength to ensure more accurate and physically meaningful predictions. In this paper, we aim to build upon the work of Zalatan et al.\yrcite{Zalatan2023,Zalatan_igarss,zalatan_icip} by incorporating the idea of physics-informed learning and proposing PSAGE-LSTM, a physics-informed graph neural network that combines the GraphSAGE framework with long short-term memory (LSTM) structure and physical node features. In our experiment, we use the thickness information captured in the Greenland region in 2012 and physical features from the Model Atmospheric Regional (MAR) weather model, aiming to predict the spatio-temporal pattern of deep ice layers based on the recently formed ice layers.

Our major contributions are: 1) We develop the novel network, PSAGE-LSTM, a physics-informed graph neural network that combines the GraphSAGE framework for graph feature learning with LSTM structure for capturing temporal changes, and introduces physical properties of ice sheets as physical node features. 2) We conducted extensive experiments on comparison with non-inductive, non-physical networks, and our proposed PSAGE-LSTM can consistently have a lower root mean squared error.

\section{Related Work}

\subsection{Internal Ice Layer Tracking}

A few automatic algorithms have been proposed to detect ice layer boundaries from raw echogram images in the past\cite{Carrer_2017, Ferro_2013,Panton_2013, Koenig_2016,MacGregor_2015}. However, a significant downside for these algorithms is the scalability for larger datasets. In recent years, deep learning techniques, especially convolutional neural networks and generative adversarial networks, have also been applied to precisely extract ice layer boundary positions from raw echogram images\cite{DeepIceLayerTracking,DeepLearningOnAirborneRadar, Rahnemoonfar_2021_JOG, Yari_2021_JSTAR}. Their results show that the major obstacles are the noise in echogram images and the lack of high-quality datasets and annotations. Some researchers also utilize the idea of physics-informed machine learning for detecting internal ice layer boundaries, where physical constraints are introduced to better denoising the raw echogram images via wavelet transform\cite{varshney2021refining,DeepHybridWavelet} or provide better initialization to neural networks\cite{LearnSnowLayerThickness}. Unlike most previous ice layer tracking methods that apply convolutional neural networks to raw echogram images, our proposed method uses a graph neural network to determine the thickness of deep ice layers, where the graph neural network is shown to be less sensitive to noise and has a more stable performance. 

\subsection{Graph Neural Network for Ice Thickness Prediction}

Zalatan et al. \yrcite{Zalatan2023,Zalatan_igarss,zalatan_icip} applied the AGCN-LSTM network to predict the thickness of deep ice layers. By combining graph convolution network (GCN) with long short-term memory (LSTM), a variant of the recurrent neural network, the GCN-LSTM network can learn the spatial relations within individual graph representations and temporal changes over time between different graphs. Zalatan et al. also use the EvolveGCNH network\cite{EGCN} as an adaptive layer, which improves the model performance by enabling the model to learn more complicated features and be more robust to noise. Compared with Zalatan et al. \yrcite{Zalatan2023,Zalatan_igarss,zalatan_icip}, our proposed method introduces physical node features that provide auxiliary information and physical constraints.

\section{Dataset}
\subsection{Raw Echogram Images}
We will use the internal ice layer dataset created based on raw echogram images captured by airborne snow radar sensors over the Greenland region. Specifically, our dataset is captured in 2012 through the snow radar sensor operated by the Center for Remote Sensing of Ice Sheets (CReSIS), as part of NASA's Operation IceBridge Mission. All the data files are available at the CReSIS website (https://data.cresis.ku.edu/). Figure \ref{fig:dataset} (a) shows an example of the captured raw echogram image. Each echogram image has 256 pixels in its width, with the depth varying from 1200 pixels to 1700 pixels. In the echogram images, the value of each pixel is determined by the corresponding signal reflection strength, where brighter pixels mean a high reflection strength\cite{Arnold_2020}. Figure \ref{fig:dataset} (b) is the binary-labeled images where the position of each ice layer is manually labeled out. The size of binary-labeled images will be the same as corresponding raw echogram images. Each pixel in the binary-labeled images can either be a layer boundary (white) or a background (black). In each echogram and its corresponding binary-labeled image, the horizontal axis represents the along-track direction, and the vertical axis represents the depth and ice layer accumulations. The thickness of ice layers can be calculated based on their upper and lower boundaries. Additionally, when capturing echogram images, the airplane records latitude and longitude simultaneously.

\subsection{Graph Data Generations}

\begin{figure*}
    \centerline
    {
        \includegraphics[width=0.9\textwidth]{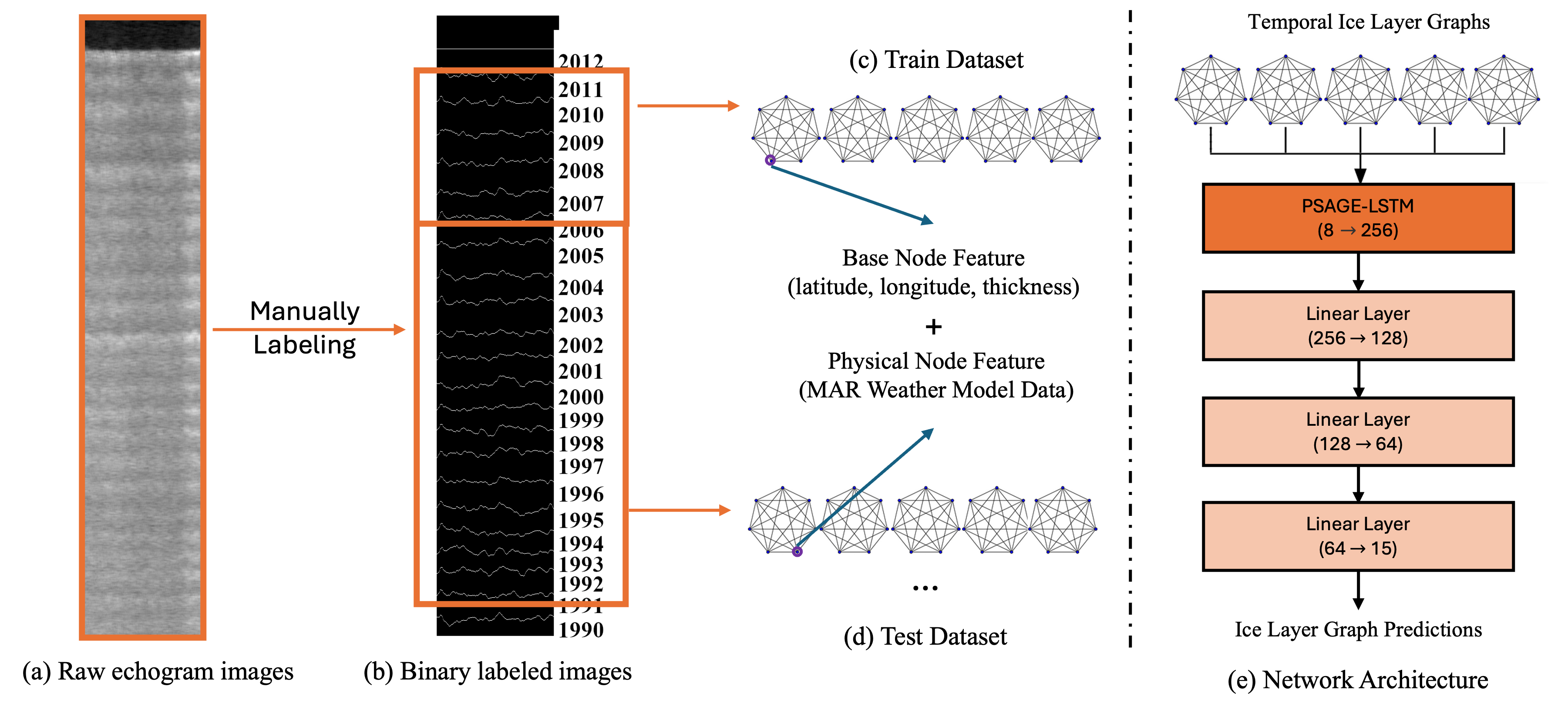}
    }
    \caption{Diagram of Our Dataset and Network. (a) Raw echogram image captured by airborne snow radar sensor. (b) Binary labeled images showing internal ice layers (c) Our training dataset is generated based on internal layers between 2007 and 2011 (d) Our test dataset is generated based on internal layers between 1992 and 2006 (e) Network Architecture of proposed PSAGE-LSTM}
    \label{fig:dataset}
\end{figure*}

In our work, we will generate our graph dataset based on the thickness information from binary-labeled images. As shown in Figure \ref{fig:dataset}, we will use the shallow five ice layers (2007-2011) to learn the spatio-temporal features of internal ice layers and provide precise predictions for the thickness of fifteen deep ice layers (1992-2006). To ensure high quality and enough valid information in our dataset, we will do some pre-processing steps and only use those binary-labeled images with at least 20 complete layers. Images may be eliminated due to insufficient internal layers or the incompleteness of layers. These pre-processing steps will reduce the number of valid images to 1660. We will split them into the training, validation, and test datasets with a ratio of 3:1:1, resulting in 996 images in the training dataset, 332 images in the validation dataset, and 334 images in the test dataset.

Graph representations will be generated based on the layer pixel value vector that shows the relative pixel position of ice layer boundaries in the image, together with each image's recorded latitude and longitude. As shown in Figure \ref{fig:dataset}, each binary-labeled image will be converted independently into a sequence of five temporal graphs as a training dataset and a sequence of fifteen temporal graphs as a test dataset, where each graph represents the ice layer formed in a particular year. Each graph will contain 256 nodes, corresponding to the 256 pixels in the width of binary-labeled images. Nodes in each graph will be fully connected and undirected. The edge weights between any node $i,j$ are calculated as:
\begin{equation}
\resizebox{.9\hsize}{!}{$w_{i,j}  = \frac{1}{2\arcsin{(hav(\phi_j-\phi_i)+\cos{\phi_i}\cos{\phi_j}hav(\lambda_j-\lambda_i))}}$}
\end{equation}where $hav(\theta) = \sin^2{(\frac{\theta}{2})}$, $\phi$ is the latitude and $\lambda$ is the longitude. Each node will have three base node features: latitude, longitude, and the thickness of the ice layer derived from the relative boundary positions.

\section{Key Designs}

Our proposed PSAGE-LSTM is a physics-informed version of the GraphSAGE-LSTM, which combines GraphSAGE with LSTM and incorporates physics as additional physical node features. In this section, we will discuss the design of GraphSAGE-LSTM, the chosen physical node features, and the network architecture of our proposed PSAGE-LSTM.

\subsection{Design of GraphSAGE-LSTM}
GraphSAGE \cite{hamilton2018inductive_graphsage} is an inductive framework that generates node embedding for unseen data based on local neighbor sampling and feature aggregation\cite{ZHOU202057_Review}. For node $i$ and its node feature $x_i$, GraphSAGE will sample from its neighbor nodes and aggregate sampled neighbor node features, defined as: 
\begin{equation}
\textbf{x}'_i = \textbf{W}_1 \textbf{x}_i + \textbf{W}_2 \cdot \text{mean}_{j \in \mathcal{N}(i)} \textbf{x}_j
\label{equation:graphsage}
\end{equation}
where $\textbf{x}'_i$ is the output of GraphSAGE network, $W_1,W_2$ are the learnable layer weights, $\mathcal{N}_i$ is the sampled neighbor list of node $i$ that includes neighbor with different depth, $\textbf{x}_j$ is the node feature of neighbors and $\text{mean}$ is the aggregator function that may be replaced with other functions. Through sampling from the neighbor nodes with different depths, GraphSAGE only learns from limited sampled neighbors instead of the complete input graph, which reduces the adverse effects of possible outliers and enhances the model's generalization ability. Compared with graph convolution, GraphSAGE with mean aggregator can be understood as a linear approximation of localized spectral convolution \cite{hamilton2018inductive_graphsage}.

As proposed by Seo et al.\yrcite{seo2016structured}, GCN-LSTM extends the LSTM structure for graph data by replacing the multiplication between model weights and inputs with graph convolution. We will combine GraphSAGE with LSTM similarly: Instead of replacing the multiplication in LSTM with GCN, we will replace it with GraphSAGE operations. An adaptive GraphSAGE-LSTM network can be built in a similar way to Zalatan et al.\yrcite{Zalatan2023,Zalatan_igarss,zalatan_icip} by adding the EvolveGCN layer as an adaptive layer.

\subsection{Physical Node Features from Model Atmospheric Regional (MAR) weather model}
Model Atmospheric Regional (MAR) is a regional weather model that can provide historical meteorological and annual climate measurements for the Greenland region\cite{MAR_2020, MAR2021}. Through proper interpolation, MAR measurements can be synchronized with snow radar data using each echogram image's latitude and longitude information, providing annual climate measurements as supplementary information. 

Our work will use physical features synchronized from MAR model data v3.10 by 2D Delaunay triangulation interpolation. We focus on five physical properties: snow mass balance, surface temperature, meltwater refreezing, height change due to melting, and snowpack heights. Experiments are conducted with different combinations of physical properties as physical node features, where some combinations may enhance the final results while some combinations of physical features will undermine them. This paper will only report the results of the best physical node feature combination.

\subsection{Architecture of PSAGE-LSTM}
Figure \ref{fig:dataset} (e) shows that our proposed network takes a sequence of 5 temporal ice layer graphs as input. Due to different combinations of physical features, the input sequence may have different numbers of node features for different experiments, ranging from 3 (only base node features) to 8 (including all the base and physical node features). To keep a constant number of input channels among all the experiments, we will set the value of those unchosen physical features to 0. The output of our proposed PSAGE-LSTM layer will have 256 channels. Features learned by PSAGE-LSTM will be passed into two hidden linear layers with 128 and 64 output channels and a last linear layer with 15 output channels for final prediction. Each output channel represents the prediction of the ice layer for one year. The hardswish activation function is used between each layer, and Dropout with $p=0.2$ is used between the three linear layers. 
\section{Experiment and Results}
\subsection{Experiment Design}
To verify the design of PSAGE-LSTM, we compare its performance with several different graph neural networks, including non-inductive, non-physical models like GCN-LSTM and AGCN-LSTM and non-physical models like GraphSAGE-LSTM and Adaptive GraphSAGE-LSTM.

All the graph neural networks are trained on 8 NVIDIA A5000 GPUs and Intel(R) Xeon(R) Gold 6430 CPU. The loss function is the mean-squared error loss. The Adam optimizer\cite{kingma2017adam} is used with an initial learning rate of 0.01 and a weight decay coefficient of 0.0001. A learning rate scheduler is used to halve the learning rate every 75 epochs. We train the GCN-LSTM and AGCN-LSTM for 300 epochs to ensure convergence. For the GraphSAGE-LSTM, Adaptive GraphSAGE-LSTM, and PSAGE-LSTM, we extend the training to 450 epochs to guarantee their convergence. Five different sets of training, validation, and test datasets are generated by applying different random permutations of the entire 1600 valid images before the train-test split. All the graph neural networks will be trained on the same five sets. 

\subsection{Results}
We calculate the root mean squared error (RMSE) for every trial between model prediction and ground-truth thickness information for ice layers formed from 1992 to 2006. The mean and standard deviation of five trials are reported as the model performance, shown in Table \ref{table:result}. By applying the physics-informed learning framework, the proposed PSAGE-LSTM network can have a significantly enhanced performance. Moreover, we also find that using EvolveGCN as an adaptive layer may not be stable for larger datasets or different models.

\begin{table}
\caption{Experiment results of GCN-LSTM, AGCN-LSTM, GraphSAGE-LSTM, Adaptive GraphSAGE-LSTM, and our proposed PSAGE-LSTM model. Results are reported as the mean and standard deviation of the RMSE on the test dataset over five individual trials.}
\label{table:result}
\vskip 0.15in
\begin{center}
\begin{small}
\begin{sc}
\begin{tabular}{cc}
\toprule
Model & Result \\
\midrule
GCN-LSTM &  3.3096 $\pm$ 0.0689 \\
AGCN-LSTM & 3.5365 $\pm$ 0.0672 \\
GraphSAGE-LSTM & 3.1872 $\pm$ 0.0511 \\
Adaptive GraphSAGE-LSTM & 3.4099 $\pm$ 0.0759 \\
PSAGE-LSTM (ours) & \textbf{2.8526 $\pm$ 0.0748} \\
\bottomrule
\end{tabular}
\end{sc}
\end{small}
\end{center}
\vskip -0.1in
\end{table}

\section{Conclusion}
In this work, we proposed PSAGE-LSTM, a physics-informed graph neural network that predicts the thickness of fifteen deep Greenland ice layers based on the thickness information of five shallow layers. Our proposed method performed consistently better than the corresponding non-physics, non-inductive models.

\bibliography{ref}
\bibliographystyle{icml2024}

\end{document}